\documentclass[final]{cvpr}

\usepackage{times}
\usepackage{epsfig}
\usepackage{graphicx}
\usepackage{amsmath}
\usepackage{amssymb}
\usepackage{booktabs,multirow}
\usepackage{array}
\usepackage{caption} 
\usepackage[pagebackref=true,breaklinks=true,colorlinks,bookmarks=false]{hyperref}

\begin{document}

\title{When Handcrafted Features and Deep Features Meet Mismatched Training and Test Sets for Deepfake Detection}

\author{Ying Xu, Sule Yildirim Yayilgan \\
\normalsize Norwegian University of Science and Technology, Gj\o vik, Norway\\
{\tt\small \{ying.xu, sule.yildirim\}@ntnu.no}
}

\maketitle

\begin{abstract}
The accelerated growth in synthetic visual media generation and manipulation has now reached the point of raising significant concerns and posing enormous intimidations towards society. There is an imperative need for automatic detection networks towards false digital content and avoid the spread of dangerous artificial information to contend with this threat. In this paper, we utilize and compare two kinds of handcrafted features(SIFT and HoG) and two kinds of deep features(Xception and CNN+RNN) for the deepfake detection task. We also check the performance of these features when there are mismatches between training sets and test sets. Evaluation is performed on the famous FaceForensics++ dataset, which contains four sub-datasets, Deepfakes, Face2Face, FaceSwap and NeuralTextures. The best results are from Xception, where the accuracy could surpass over 99\% when the training and test set are both from the same sub-dataset. In comparison, the results drop dramatically when the training set mismatches the test set. This phenomenon reveals the challenge of creating a universal deepfake detection system.
\end{abstract}

\section{Introduction}
The human face often works as the most critical part of our bodies not only because it presents the first impression to others but also for the security matters. For instance, faces have much information about  personal data of people, such as age and gender. Face images are widely used in various fields including verification of identity, face payment and track down criminals. With the assistance of the Internet, smartphones and social media, countless face media appear in public every day. Provided the convenience of accessing large amounts of data, creating natural, manipulated media assets may be easy. There are certainly entertaining and non-nefarious face manipulation applications, such as movie productions, photography, video games and virtual reality. Meanwhile, concerns have been raised about possible improper uses, which would cause severe trust issues and security concerns in our society. Therefore, a universal deepfake detection system is a urgent need.

Nowadays, face manipulation techniques are not limited to manual image editing software like Adobe PhotoShop \footnote{https://www.adobe.com/products/photoshopfamily.html} or GIMP \footnote{https://www.gimp.org/} for instance. We have gone far beyond these elementary manipulations to adding mythical figures and inserting or removing objects. Some easy-to-use applications \footnote{https://github.com/deepfakes/faceswap} \footnote{https://github.com/iperov/DeepFaceLab}, which only require little training process, can be operated in Graphical User Interfaces(GUIs). Fired by the recent achievement of Generative Adversarial Networks (GANs) along with the availability of GPUs, it is simple for an amateur user to produce completely artificial yet hyperrealistic contents. The realism achieved by the machine learning components is too high to assess whether a face picture is naturally captured by a camera or artificially produced for human eyes. However, if we observe the generated videos or images closely, we could still find some artifacts. For example, in GAN-generated faces, a mismatch may happen to different asymmetry forms, such as a missing earring on one ear or the different colors between two eyes, unconvincing specular reflections, abstaining or represented as white blobs or roughly modeled teeth.

\begin{figure}[h]
  \includegraphics[width=\linewidth]{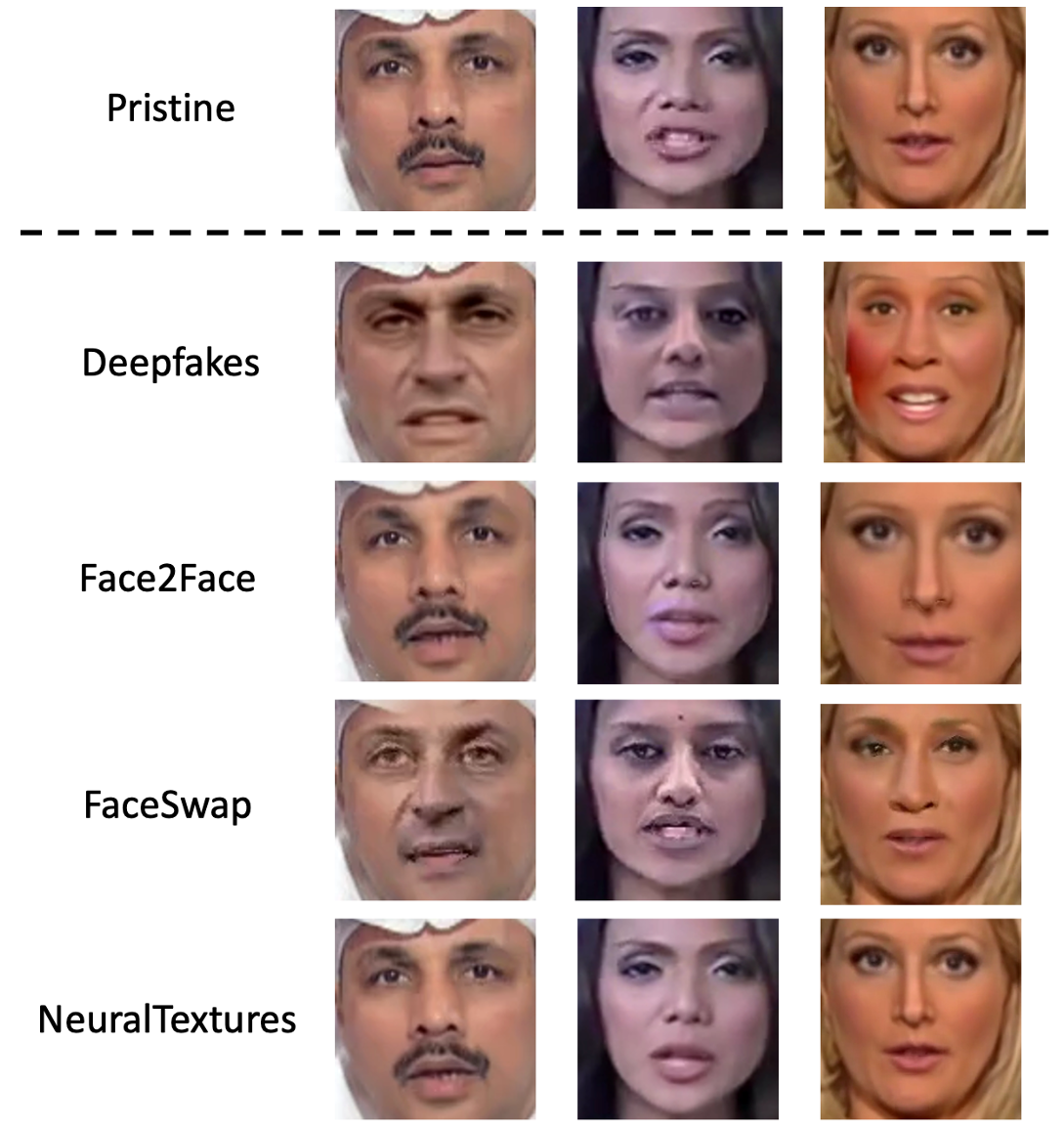}
  \caption{Face images from different sub-datasets. The first row is  Pristine frames. The second, third, fourth and fifith row are from Deepfakes, Face2Face, FaceSwap and NeuralTextures respectively.}
  \label{show dataset}
\end{figure}

The dataset we are using in this work is FaceForensics++ consisting of 1000 original videos. All of the one thousand sequences have been manipulated by Deepfakes, Face2Face, FaceSwap and NeuralTextures respectively, as prominent representatives for facial manipulations at random compression level and size. Figure \ref{show dataset} shows the cropped faces from those five sub-datasets. For the Deepfakes dataset, faceswap Github implementation \footnote{https://github.com/deepfakes/faceswap} is applied. Two autoencoders with a shared encoder are trained to reconstruct the source's training images and the target face, respectively. After training, the trained encoder and decoder from the source face are applied to the target face to create a fake image. Face2Face \cite{thies2016face2face} is a facial reenactment system that transfers the expressions of a source video to a target video while maintaining the target person's identity. Facial expressions of both source and target videos are tracked using a dense photometric consistency measure. Reenactment is then achieved by deformation transfer between source and target. The synthesized target face re-render on top of the corresponding video stream makes the videos more realistic with the real-world illumination followed by mouth interior refinement. FaceSwap is a graphics-based approach to transfer the face region from a source video to a target video. The facial area and landmarks extracted are fed to a 3D template model. The vertices of that model projected to the image space will be the texture coordinates. This model is then back projected to the target image by minimizing the difference between the projected shape and the localized landmarks using the textures of the input image. Finally, the rendered model is blended with the image and color correction is applied. FaceSwap videos are generated by Github code \footnote{https://github.com/MarekKowalski/FaceSwap}. NeuralTextures videos are generated by using the face model to track and render corresponding UV masks. Those masks are then fed into an encoder-decoder architecture, which is optimized using Neural Textures \cite{thies2019deferred}. This network uses the original video data to learn the target person's neural texture to be trained with a photometric reconstruction loss combined with an adversarial loss.
FF++ dataset provides a benchmark with 1000 hidden images manipulated by these four methods.\footnote{http://kaldir.vc.in.tum.de/faceforensics\_benchmark/}

In this work, we applied Scale Invariant Feature Transform(SIFT) \cite{lowe2004distinctive} and Histograms of Oriented Gradients(HOG) \cite{dalal2005histograms} to extract handcrafted features. Then we feed these features to a support vector machine (SVM) for binary classification. We selected Xception \cite{chollet2017xception} and CNN+RNN as deep features extractors.We also conduct experiments to see how much performance drop for handcrafted and deep learning features if the training set and test set have a mismatch. The schemes for all the experiments are shown in Figure \ref{pipline}.

\section{Related Work}
In this section, we first look into trending detection methods. Afterwards, we summarize the progress that has been done on the FF++ dataset so far.

\subsection{General methods}
Several types of networks have been using for deepfake detection over the last few years. A general approach is to detect visible artifacts in the face. Methods following this approach highlight specific failures in the generation process that do not correctly reproduce real face details. For example, in GAN-generated faces, mismatches may happen to different asymmetry forms, such as missing earring on one ear or the different colors between two eyes. While Deepfakes often present unconvincing specular reflections or missing one, abstaining or represented as white blobs or roughly modeled teeth. The general methodologies usually detect the lack of physiological signals intrinsic to human beings that are not well captured in the synthesized videos. Therefore, some papers\cite{li2018ictu, matern2019exploiting} focused on detailed artifacts of sense organs, especially on eyes and teeth. Some papers \cite{agarwal2019protecting, yang2019exposing} choose to emphasize head poses because synthesis algorithms can produce a face with an outstanding level of realism and with many details lack a specific constraint over the locations of these components in a face. While quite a few papers \cite{li2018exposing, li2020face} choose to detect inconsistency artifacts as most of the current face manipulation methods share the common step of blending an altered face into an existing background image, which leaves intrinsic image discrepancies and wrapping artifacts across the blending boundaries. Other papers \cite{mccloskey2019detecting, li2018detection, koopman2018detection}, on the other hand, focus on the color discrepancy. 
The most significant advantage of deep neural networks is that it is unnecessary to search underlying features for detecting fake images. The networks would learn how to distinguish fake from real by themselves. By using the mature convolutional neural network (CNN), this job has become easy. A bunch of papers \cite{marra2018detection, marra2019gans, yu2019attributing, roessler2019faceforensicspp,  do2018forensics, nguyen2019use, ding2020swapped, bonettini2020video} has done experiments with CNN-based methods. Simultaneously, more reliable results \cite{guera2018deepfake, sabir2019recurrent, stehouwer2019detection} can be foreseen by approaches that take explicitly into account the temporal direction. Two-stream networks \cite{zhou2017two, chen2019attention, songsri2019complement, wu2020sstnet}, combining two different kinds of features for detection and classification tasks, are also gaining more popularity on this topic. Other focuses have been put on frequency domain \cite{durall2019unmasking, chen2020manipulated, qian2020thinking}, as well as GAN fingerprints \cite{zhang2019detecting, liu2020global, wang2020cnn, guarnera2020deepfake, guarnera2020fighting, barni2020cnn}.

\subsection{Benchmark on FF++}
FF++ dataset group is offering a benchmark for facial manipulation detection on the presence of compression based on their four manipulation methods(Deepfakes, Face2Face, FaceSwap, NeuralTextures), appended with pristine ones. This hidden dataset contains 1000 images and is welcoming everyone to give it a try. 
The author of FF++ published the dataset along with their results \cite{roessler2019faceforensicspp}. They first conducted a user study with 204 participants as a baseline of human observers. According to these observers, humans can distinguish fake videos from deepfakes with 0.782 accuracy, which is the best performance among the five sub-datasets. With an average accuracy of 0.308, it is pretty hard to tell which videos are fake in NeuralTextures. 
\par In Table \ref{table:benchmark}, we recorded all the testing results validated by the hidden dataset within the published papers in this benchmark. Dogonadze, N. et al.\cite{dogonadze2020deep} achieved the best performance in DF with a 0.936 accuracy, FS with a 0.903 accuracy, and NT with a 0.820 accuracy. They utilized Inception ResNet V1 \cite{zhang2018residual} which was pretrained on the VGGFace2 \cite{cao2018vggface2} face recognition task and revealed the importance of pretraining. Xception has a 0.839 accuracy in FS and 0.903 in FS which are still the top ones in published works.

\begin{table}[h]
\centering
\begin{tabular}{l|p{0.7cm}<{\centering}|c<{\centering}|c|c|c}
\hline
& DF & F2F & FS & NT & P   \\ \hline \hline
Human\cite{roessler2019faceforensicspp} & 0.782 & 0.502 & 0.758 & 0.308 & \textbf{0.754} \\
D. N. et al.\cite{dogonadze2020deep}&  \textbf{0.936} & 0.839 & \textbf{0.903} & \textbf{0.82} & 0.750 \\
Xception\cite{roessler2019faceforensicspp}&  0.964&  \textbf{0.869} &  \textbf{0.903} & 0.807 & 0.524 \\
MesoNet\cite{afchar2018mesonet} & 0.873 & 0.562 & 0.612 & 0.407 & 0.726 \\
X. full\cite{roessler2019faceforensicspp} & 0.745 & 0.759 & 0.709 & 0.733 & 0.510 \\
Jordi M. C.\cite{master2020project} & 0.791 & 0.730 & 0.816 & 0.72 & 0.478 \\
B. B. et al.\cite{bayar2016deep} & 0.845 & 0.737 & 0.825 & 0.707 & 0.462 \\
R. N. et al.\cite{rahmouni2017distinguishing} & 0.855 & 0.642 & 0.563 & 0.607 & 0.5 \\
C. D. et al.\cite{cozzolino2017recasting} & 0.855 & 0.679 & 0.738 & 0.78 & 0.344 \\
F. J. et al.\cite{fridrich2012rich} & 0.736 & 0.737 & 0.689 & 0.633 & 0.340\\
\hline
\end{tabular}
\captionsetup[table]{skip=3pt}
\caption{Benchmark results with published papers}
\label{table:benchmark}
\end{table}
We just listed the result with the published papers in Table \ref{table:benchmark}. There are better results with anonymous participants. The best performances for the sub-datasets are DF-1.000(ZAntiFakeBio, Leo, Aquarius, InTeLe\_), F2F-0.942(DFirt), FS-0.981(faceClassify1), NT-0.953(Two-stream-SRM-RGB1), and pristine-0.952(Cancer) according to the benchmark. It is worth noting that no method could take the first price of two sub-datasets, which indicates the difficulties of creating a universal deepfake detection network.
\section{Proposed comparison}
\subsection{Dataset}
We use FaceForensics++\cite{roessler2019faceforensicspp} as the dataset for the comparison. In this dataset, 1000 original video sequences are manipulated by four automated face manipulation methods: Deepfakes, Face2Face, FaceSwap and NeuralTextures. We chose high-quality videos with light compression (c23) in these datasets for all the experiments. The first 750 videos were used for the training and the last 250 videos were used for the test.

\begin{figure*}[h]
  \includegraphics[width=\linewidth]{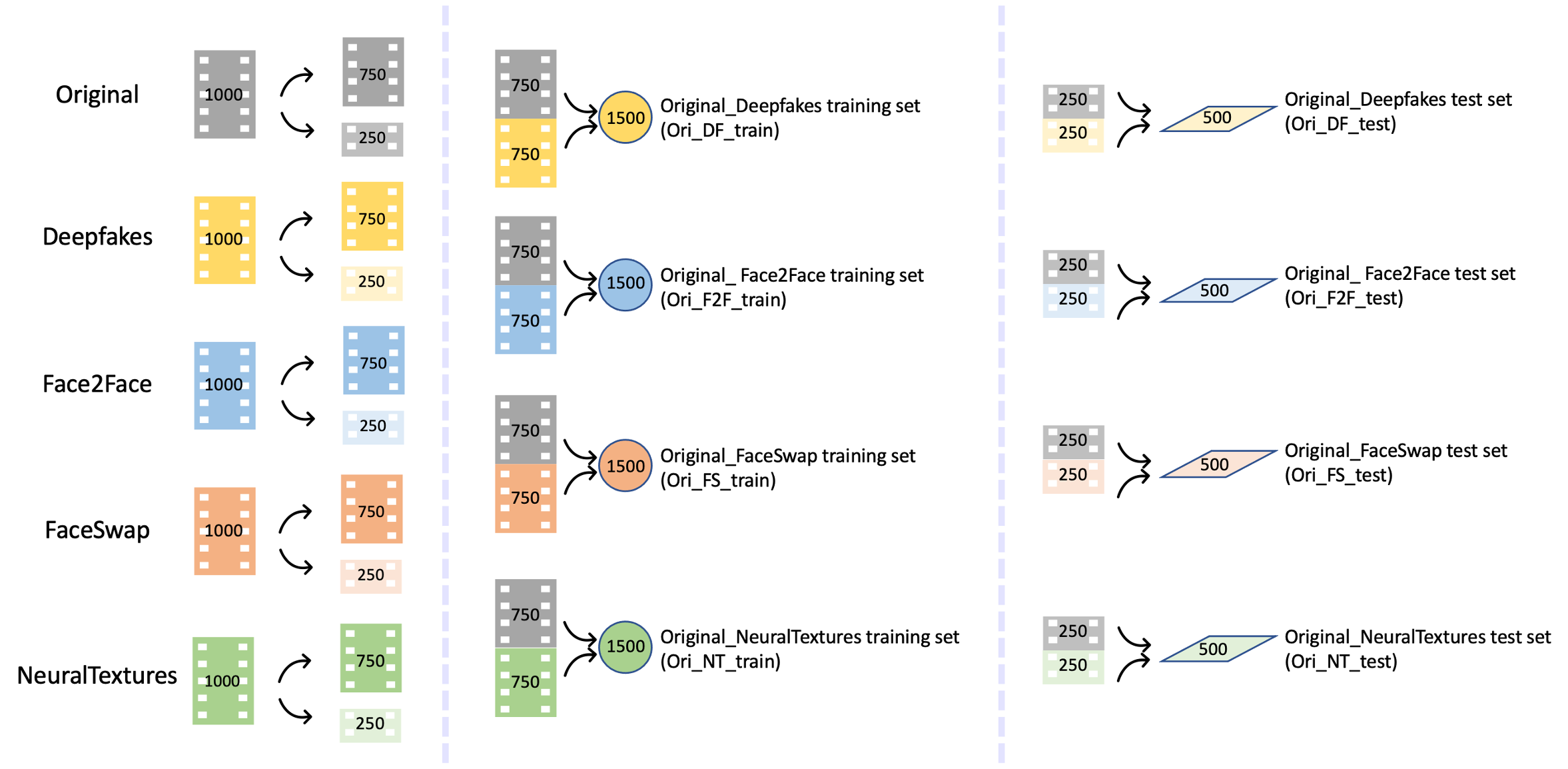}
  \caption{Dataset distribution for all five datasets in FF++. After distribution, we get eight sub-datasets. Original\_Deepfakes training set(Ori\_DF\_Train), Original\_Face2Face training set(Ori\_F2F\_Train), Original\_FaceSwap training set(Ori\_FS\_Train), and Original\_NeuralTextures training set(Ori\_NT\_Train) are for training. Original\_Deepfakes\_test set(Ori\_DF\_Test) ,Original\_Face2Face test set(Ori\_F2F\_Test), Original\_FaceSwap test set(Ori\_FS\_Test), and Original\_NeuralTextures test set(Ori\_NT\_Test) are for test.}
  \label{dataset distribution}
\end{figure*}

\par\noindent{\textbf{Dataset distribution}} The distribution of all the datasets are shown in Figure \ref{dataset distribution}. There are a total of 5,000 videos and five sub-datasets (Original, Deepfakes, Face2Face, FaceSwap, NeuralTextures)  in the FF++ dataset. Each subset owns 1,000 videos respectively. Therefore, 1,000 videos are real and 4,000 videos are fake. We first separated each subset into 750 and 250 videos. Let us call them Original-750, Original-250, Deepfakes-750, Deepfakes-250, Face2Face-750, Face2Face-250, FaceSwap-750, FaceSwap-250, NeuralTextures-750, and NeuralTextures-250. Then we combined Original-750 and Deepfakes-750 as Original\_Deepfakes training set, Original-250 and Deepfakes-250 as Original\_Deepfakes test set. The same procedure may be easily adapted to obtain Original\_Face2Face training/test set, Original\_FaceSwap training/test set and Original\_NeuralTextures training/test set.

\subsection{Facial area preprocessing}
We trained our networks only with face regions of each video after conducting facial area cropping. Since R\"{o}ssler  provided facial information for each video\footnote {https://github.com/ondyari/FaceForensicsDataProcessing}, we completed the face region cropping by extracting each frame from each video and using the bounding boxed to crop the face area.

\subsection{Networks}
In our experiments, we chose four feature extraction algorithms. Two of them, Scale Invariant Feature Transform(SIFT) and Histogram of oriented gradients(HOG), are handcrafted feature methods. The other two, Xception and CNN+RNN(Resnet-152+LSTM specifically), are deep learning networks.

\par\noindent{\textbf{SIFT}} represents Scale-Invariant Feature Transform and was first presented in \cite{lowe1999object}. After applying SIFT to images, we can get descriptors for feature keypoints. By classifying these descriptors, we can then distinguish the real and fake images.

\par\noindent{\textbf{HoG}} stands for Histogram of oriented gradients. This method became widespread in 2005 when the authors in \cite{dalal2005histograms} presented their supplemental work on HOG descriptors on the original MIT pedestrian database. After resizing an image, the gradient, magnitude and orientation of each pixel are calculated. Then this image is divided into many parts, which are decided by parameters. A histogram shows the frequency distribution of a set of continuous data will be plotted for localized divisions of an image by counting occurrences of gradient orientation. 

\par\noindent{\textbf{Xception}} is a convolutional neural network architecture based entirely on depthwise separable convolution layers \cite{chollet2017xception}. It has 36 convolutional layers forming the feature extraction base of the network. All of these layers are structured into 14 modules, all of which have a linear residual connections around them, except for the first and last modules. Although Xception is a really deep convolutional neural network, the training speed is relatively fast because of depthwise separable convolution.

\par\noindent{\textbf{ResNet-152+LSTM}} is inspired by the recurrent convolutional strategies proposed in \cite{sabir2019recurrent}. At the first stage, we use the pre-trained Resnet-152 \cite{he2016deep} to extract features. The core idea exploited in the Resnet model, residual connections, improves gradient flow significantly, allowing for much deeper models with even hundreds of layers. The author also provided ample experimental evidence showing that these residual networks are easier to optimize. At the second stage, we feed former extracted features into Long Short-Term Memory(LSTM) \cite{sak2014long}. A standard LSTM unit comprises a cell, an input gate, an output gate, and a forget gate. These gates can learn which data in a sequence is necessary to keep or useless to throw away. Then it can pass relevant information down the long chain of sequences to make predictions.

\begin{figure*}
  \includegraphics[width=\linewidth]{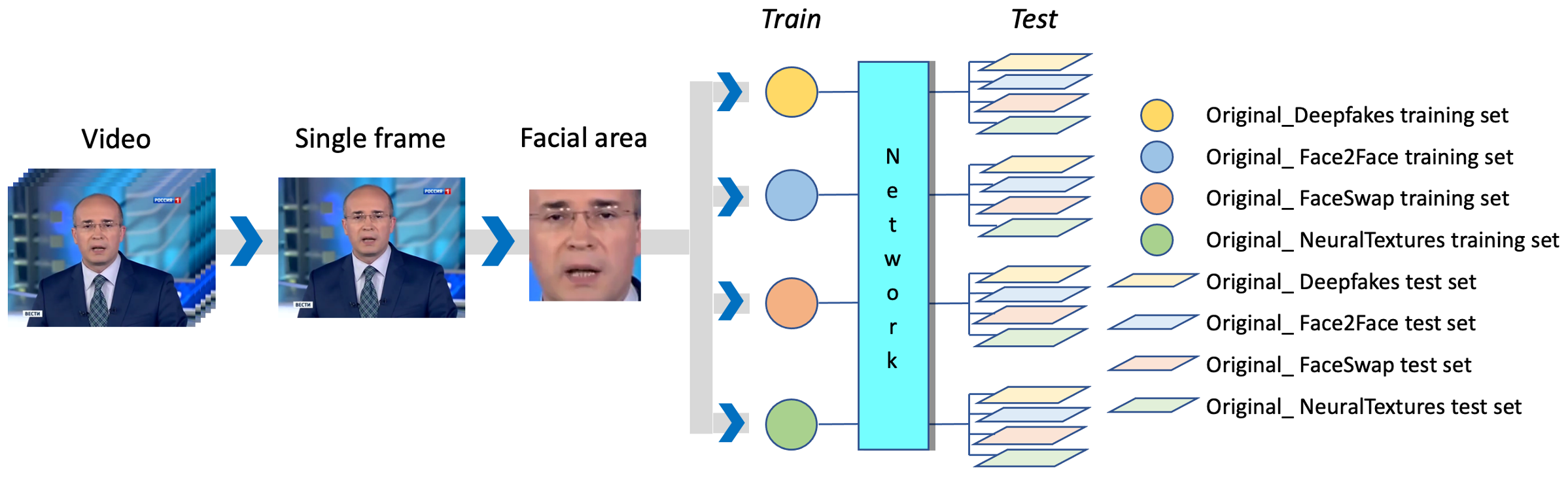}
  \caption{The pipeline of the training and test procedure. Given a video, after facial area preprocessing, we get the faces of each frame. Then we feed four training sets into the network respectively. Finally, we test each trained network with four test sets.}
  \label{pipline}
\end{figure*}

\subsection{Pipeline}
The pipeline of our training and test is demonstrated in Figure \ref{pipline}. After the facial area pre-processing, we obtain face images for each frame in all the 5000 videos. We then feed every network, SIFT, HoG, Xception, and Resnet-152+LSTM, by Original\_Deepfakes, Original\_Face2Face, Original\_FaceSwap, Original\_NeuralTextures training set. We next test the trained network by Original\_Deepfakes, Original\_Face2Face, Original\_FaceSwap, Original\_NeuralTextures test sets to the performance of these four networks and see how mismatches between training and test datasets would affect the detection results.

\section{Experiments}
\subsection{Implementation details}
All our results were genereated on an NVIDIA Grid V100D-8Q GPU.
As a limitation of time and GPU space, we have not finished SIFT and HoG experiments with the full training and test dataset. We choose 10,000 frames from each training set and 5,000 frames from each test set randomly for the experiments.

For the ResNet-152+LSTM network, we used ResNet-152 pre-trained on ImageNet without the last classifier layer. We chose the numbers 3 and 1024 for hidden layers and hidden nodes respectively in LSTM. We used a pre-trained Xception model of ImageNet as initial weights and slightly trained by our datasets. We used the cross-entropy loss and Adam optimizer with $\beta_1$ = 0.9, $\beta_2$ = 0.999, and a learning rate of 0.001. Due to GPU space, we set the batch size as 16. 

\subsection{Evaluation metrics}
We chose five evaluation metrics for our evaluation: accuracy, precision, sensitivity, F1 and F2 score.

\begin{itemize}
    \item \textbf{Accuracy} is the quintessential classification metric. It is the fraction of true results among the total number of cases examined. The formula for accuracy is $\frac{TP+TN}{TP+TN+FP+FN}$, where TP, TN, FP, and FN are true positive, true negative, false positive, and false negative, respectively.
    \item \textbf{Precision} is the portion of relevant instances among the retrieved instances. The formula for precision is $\frac{TP}{TP+FP}$.
    \item \textbf{Recall}, also known as sensitivity, is the fraction of the total amount of retrieved relevant instances. The formula for sensitivity is $\frac{TP}{TP+FN}$.
    \item \textbf{F-score} is an essential measure for binary classification. The general formula for it is $F_\beta=(1+\beta^2) \cdot \frac{precision \cdot recall}{(\beta^2 \cdot precision) + recall}$. When $\beta$ equals 1, it turns out to be F1 score. The result would be the harmonic mean of precision and recall. When $\beta$ equals 2, the recall weighs higher than precision. F2 score shows the ability of a network that classifies as many positive samples as possible, rather than maximizing the number of correct classifications.
\end{itemize}

\subsection{Results}
\begin{table*}[ht]
  \centering
  \renewcommand{\arraystretch}{1.1}
\begin{tabular*}{0.91\linewidth}{p{2cm}<{\centering}|p{1.9cm}|p{1.7cm}|p{1.3cm}<{\centering}|p{1.3cm}<{\centering}|p{1.3cm}<{\centering}|p{1.3cm}<{\centering}|p{1.6cm}<{\centering}}
\hline 
    \textbf{Model} & \textbf{Train set} & \textbf{Test set} & \textbf{Accuracy} & \textbf{Precision} & \textbf{Recall} & \textbf{F1 score} & \textbf{F2 score}\\ \hline \hline
    \multirow{16}{*}{SIFT + SVC} 
        & \multirow{4}{*}{Ori\_DF\_train} & Ori\_DF\_test & 61.48 & 78.41 & 29.74 & 43.12 & 33.95 \\
        &    & Ori\_F2F\_test & 50.20 & 56.12 & 8.32 & 14.49 & 10.03 \\
        &    & Ori\_FS\_test & 54.56 & 44.90 & 7.29 & 12.55 & 8.76 \\
        &    & Ori\_NT\_test & 55.84 & 55.48 & 10.77 & 18.04 & 12.84 \\ \cline{2-8}
        & \multirow{4}{*}{Ori\_F2F\_train} & Ori\_DF\_test & 51.30 & 50.66 & 31.16 & 38.59 & 33.76 \\
        &    & Ori\_F2F\_test & 60.16 & 63.89 & 49.33 & 55.67 & 51.69 \\
        &    & Ori\_FS\_test & 53.06 & 46.17 & 30.25 & 36.55 & 32.49 \\
        &    & Ori\_NT\_test & 52.22 & 45.32 & 28.55 & 35.03 & 30.83 \\ \cline{2-8}
        & \multirow{4}{*}{Ori\_FS\_train} & Ori\_DF\_test & 51.30 & 50.66 & 31.16 & 38.59 & 33.76 \\
        &    & Ori\_F2F\_test & 60.16 & 63.89 & 49.33 & 55.67 & 51.69 \\
        &    & Ori\_FS\_test & 53.06 & 46.17 & 30.25 & 36.55 & 32.49 \\
        &    & Ori\_NT\_test & 52.22 & 45.32 & 28.55 & 35.03 & 30.83 \\ \cline{2-8}
        & \multirow{4}{*}{Ori\_FS\_train} & Ori\_DF\_test & 53.00 & 56.88 & 17.68 & 26.97 & 20.50 \\
        &    & Ori\_F2F\_test & 51.26 & 56.98 & 15.93 & 24.90 & 18.61 \\
        &    & Ori\_FS\_test & 52.64 & 38.15 & 9.57 & 15.31 & 11.26 \\
        &    & Ori\_NT\_test & 59.44 & 63.23 & 24.16 & 34.96 & 27.56 \\ \hline
    \multirow{16}{*}{HoG + SVC} 
        & \multirow{4}{*}{Ori\_DF\_train} & Ori\_DF\_test & \textbf{74.26} & 65.74 & 78.35 & 71.50 & 75.46 \\
        &    & Ori\_F2F\_test & 51.78 & 22.36 & 56.19 & 31.99 & 43.14 \\
        &    & Ori\_FS\_test & 51.44 & 14.05 & 38.25 & 20.55 & 28.45 \\
        &    & Ori\_NT\_test & 56.54 & 25.75 & 53.85 & 34.84 & 44.20 \\ \cline{2-8}
        & \multirow{4}{*}{Ori\_F2F\_train} & Ori\_DF\_test & 49.82 & 27.21 & 48.06 & 34.75 & 41.67 \\
        &    & Ori\_F2F\_test & 71.80 & 71.10 & 72.70 & 71.89 & 72.37 \\
        &    & Ori\_FS\_test & 56.00 & 34.77 & 51.15 & 41.40 & 46.75 \\
        &    & Ori\_NT\_test & 56.28 & 36.26 & 52.23 & 42.80 & 48.00 \\ \cline{2-8}
        & \multirow{4}{*}{Ori\_FS\_train} & Ori\_DF\_test & 46.84 & 14.91 & 39.14 & 21.59 & 29.54 \\
        &    & Ori\_F2F\_test & 51.66 & 27.13 & 54.73 & 36.28 & 45.48 \\
        &    & Ori\_FS\_test & 73.60 & 68.05 & 71.51 & 69.74 & 70.79 \\
        &    & Ori\_NT\_test & 52.36 & 21.81 & 44.32 & 29.23 & 36.74 \\ \cline{2-8}
        & \multirow{4}{*}{Ori\_FS\_train} & Ori\_DF\_test & 58.14 & 45.78 & 59.60 & 51.79 & 56.21 \\
        &    & Ori\_F2F\_test & 56.96 & 44.05 & 60.38 & 50.93 & 56.21 \\
        &    & Ori\_FS\_test & 50.94 & 28.05 & 42.60 & 33.83 & 38.59 \\
        &    & Ori\_NT\_test & 64.80 & 59.49 & 61.33 & 60.40 & 60.96 \\ 
    \lasthline
    \end{tabular*}
  \caption{Classification results on different test sets using SIFT+SVC and HoG+SVC}
  \label{table:hand result}
\end{table*}

\begin{table*}[ht]
  \centering
  \renewcommand{\arraystretch}{1.1}
  \newcommand{\tabincell}[2]{\begin{tabular}{@{}#1@{}}#2\end{tabular}}
\begin{tabular*}{0.85\linewidth}{p{2cm}<{\centering}|p{1.4cm}|p{1.4cm}|p{1.3cm}<{\centering}|p{1.3cm}<{\centering}|p{1.3cm}<{\centering}|p{1.3cm}<{\centering}|p{1.3cm}<{\centering}}
\hline 
    \textbf{Model} & \textbf{Train set} & \textbf{Test set} & \textbf{Accuracy} & \textbf{Precision} & \textbf{Recall} & \textbf{F1 score} & \textbf{F2 score}\\ \hline \hline
    \multirow{16}{*}{Xception} 
        & \multirow{4}{*}{Ori\_DF} & Ori\_DF & \textbf{98.77} & \textbf{99.28} & \textbf{98.23} & \textbf{98.75} & \textbf{99.06} \\
        &    & Ori\_F2F & 51.10 & 83.72 & 3.57 & 6.85 & 15.25 \\
        &    & Ori\_FS & 55.05 & 8.33 & 0.08 & 0.16 & 0.38 \\
        &    & Ori\_NT & 57.34 & 85.69 & 5.23 & 9.86 & 21.02 \\ \cline{2-8}
        & \multirow{4}{*}{Ori\_F2F} & Ori\_DF & 52.55 & 4.68 & 91.77 & 8.91 & 19.43 \\
        &    & Ori\_F2F & \textbf{99.36} & \textbf{99.14} & \textbf{99.59} & \textbf{99.37} & \textbf{99.50} \\
        &    & Ori\_FS & 55.72 & 1.22 & 70.43 & 2.40 & 5.71 \\
        &    & Ori\_NT & 55.65 & 1.09 & 67.98 & 2.14 & 5.11 \\ \cline{2-8}
        & \multirow{4}{*}{Ori\_FS} & Ori\_DF & 50.24 & 0.08 & 13.31 & 0.15 & 0.37 \\
        &    & Ori\_F2F & 50.13 & 1.42 & 74.97 & 2.79 & 6.60 \\
        &    & Ori\_FS & \textbf{99.43} & \textbf{99.33} & \textbf{99.40} & \textbf{99.37} & \textbf{99.39} \\
        &    & Ori\_NT & 55.17 & 0.09 & 13.31 & 0.18 & 0.45 \\ \cline{2-8}
        & \multirow{4}{*}{Ori\_NT} & Ori\_DF & 74.80 & 93.88 & 52.59 & 67.41 & 57.66 \\
        &    & Ori\_F2F & 53.45 & 76.55 & 10.86 & 19.02 & 13.11 \\
        &    & Ori\_FS & 54.03 & 21.12 & 1.12 & 2.13 & 1.38 \\
        &    & Ori\_NT & \textbf{93.61} & \textbf{95.55} & \textbf{89.86} & \textbf{92.62} & \textbf{90.94} \\ \hline
    \multirow{16}{*}{CNN+RNN}
        & \multirow{4}{*}{Ori\_DF} & Ori\_DF & \textbf{88.53} & 89.88 & 86.60 & 88.21 & 87.24 \\
        &    & Ori\_F2F & 61.24 & 77.45 & 32.45 & 45.74 & 36.72 \\
        &    & Ori\_FS & 51.79 & 24.26 & 3.81 & 6.59 & 4.58 \\
        &    & Ori\_NT & 68.42 & 77.55 & 41.10 & 53.73 & 45.37 \\ \cline{2-8}
        & \multirow{4}{*}{Ori\_F2F} & Ori\_DF & 63.42 & 75.53 & 38.75 & 51.22 & 42.93 \\
        &    & Ori\_F2F & 85.47 & 87.26 & 83.30 & 85.24 & 84.07 \\
        &    & Ori\_FS & 57.43 & 56.46 & 19.87 & 29.39 & 22.82 \\
        &    & Ori\_NT & 65.52 & 71.28 & 38.02 & 49.59 & 41.93 \\ \cline{2-8}
        & \multirow{4}{*}{Ori\_FS} & Ori\_DF & 48.05 & 38.64 & 8.20 & 13.52 & 9.73 \\
        &    & Ori\_F2F & 55.05 & 64.91 & 23.33 & 34.33 & 26.76 \\
        &    & Ori\_FS & 86.34 & 84.30 & 85.25 & 84.77 & 85.06 \\
        &    & Ori\_NT & 52.84 & 39.01 & 10.16 & 16.12 & 11.92 \\ \cline{2-8}
        & \multirow{4}{*}{Ori\_NT} & Ori\_DF & 69.21 & 80.23 & 50.26 & 61.81 & 54.32 \\
        &    & Ori\_F2F & 64.44 & 77.52 & 41.37 & 53.95 & 45.63 \\
        &    & Ori\_FS & 52.58 & 36.80 & 8.80 & 14.20 & 10.38 \\
        &    & Ori\_NT & 84.12 & 84.03 & 79.52 & 81.71 & 80.38 \\ 
    \lasthline
    \end{tabular*}
  \caption{Classification results on different test sets using Xception and CNN+RNN}
  \label{table:deep result}
\end{table*}

Table \ref{table:hand result} shows the classification results for handcrafted features. HoG performs better than SIFT in general.  However, they do not have a high accuracy even when the training set and test set are from the same dataset, indicating that handcrafted features are not suitable for fake face detection tasks. We suppose the main reason is that handcrafted features, or SIFT and HoG, are mainly used for object detection or landmarks matching. They could perform poorly when the task regards details. In deepfake detection tasks, the inputs are always faces, no matter real or fake. Therefore, it may be unachievable to use handcrafted features to detect deepfakes.

Table \ref{table:deep result} shows the results for deep features. The advantage of deep learning methods is that we can get high performance if the training set and the test set matches. The Xception experimental results of DF, F2F, FS are nearly perfect. However, there is a slight drop in NT, where accuracy and F1-score are 93.61\% and 92.96\%. One possible reason is that the photo-realistic rendering effect of Neural Textures is high. We could observe this from Figure \ref{show dataset} easily. It made F2F videos much more genuine so that those videos could escape from the detection of a deep network. The Resnet-152+LSTM network does not perform as well as Xception. The best accuracy, 88.53\%, is from DF. The modest performance may be caused by LSTM taking only the information from previous pixels within one image rather than previous frames within on video. Surprisingly, Xception performs poorly when training set and test set mismatches. Most of the recalls tested on the inconsistent test set are less than 5.00\%. It shows that this network does not work at all on new datasets. The average accuracy for mismatch test of Resnet-152+LSTM is around 55\%. Except for three low recalls under 10\%, the rest average is 34\%, which is higher than Xception but unusable.

The recall of handcrafted features is generally higher than deep features. We assume the reason would be that deep features are too specific when classifying, so it only suits videos from the same generation network. While handcrafted features are more inclusive, but it is more like a random guess.

\section{Limitation and future work}
Due to the space of GPU and time limitation for implementation, we only use 10000 frames randomly chosen from all the training sub-dataset and the first ten SIFT descriptors of each image for training. Besides, we fed the SVM with one SIFT descriptor in one image respectively, instead of keeping all descriptors from one frame in one input. Training with one SIFT descriptor means all the descriptors from one frame have the same label no matter the specific descriptor is real or fake. Although we used the facial area descriptor, we could not guarantee that every descriptor in a fake video frame is not real. However, if we combine all descriptors from one frame as one input, the trained network may have some pattern to detect a fake frame, which is not reasonable either. 
\par We only used the confusion matrix to calculate the accuracy, precision, recall, and F-scores. However, the ROC curve (Receiver Operating Characteristic curve) and the AUC score(Area Under the ROC Curve) would better explain the results. By observing the ROC curve, we could recognize the tradeoff between sensitivity and specificity for all possible thresholds rather than just the one chosen by the modeling technique. In the future, we would plot the ROC curve and calculate the AUC score for our results.
\par We made full use of the spatial data of the FF++ dataset, but we lost the temporal information within videos. With limited time, we could only do experiments with a single image as an input. There are two options if we want to take temporal data into account. One is to create new videos using the cropped facial area we made previously, which needs to deal with the position changes of the head and requires head alignment. The other is to use the video as an input and recognize and crop the face area dynamically, demanding more GPU space and advanced coding skills. The latter one is the final goal, to make an end-to-end universal deepfake detection network.
\par We could do more experiments to make the results more abundant and complete. For example, we could create a two-stream network where we could combine handcrafted and deep features to make predictions. We could also combine two kinds of deep features for fusion prediction. Additionally, the portion between training and test could be changed. What we separate now is 75\% of the FF++ dataset for the training set and 25\% for the test set. In the training set, 80\% for training and 20\% for validation. So the portion among training, validation and test is 60\%, 15\% and 25\%. In the future, we could try 70\%, 15\% and 15\% or 80\%, 10\% and 10\% and see some networks could perform better. Besides, we only did experiments on the c23 videos rather than raw and c40(highly compressed) videos. We hope it would be enough space for the middle products and it will be challenging to process the raw videos with limited GPU RAM. Finally, we assume that there is more space to improve the results by fine tuning the parameters.

\section{Conclusion}
In this work, we utilized SIFT+SVC, HoG+SVC, Xception, and CNN+RNN to detect deepfakes in FF++ dataset. In the meantime, we have tested how networks would perform when the training set mismatches with the test set. The performances of handcrafted features could only get the highest 74.26\% accuracy, which drops moderately to 56\% averagely when mismatches between training and test sets. While deep features work perfectly and could reach over 99\% if the data are from the same sub-dataset, the accuracy drops dramatically when training sets and test sets mismatch. Our goal is to create a universal end-to-end deepfake detection system that would perform well on videos produced by every generation method.

{\small
\bibliographystyle{ieee_fullname}
\bibliography{egbib}
}

\end{document}